%% file: IEEE-conference-template-062824.tex
\documentclass[conference]{IEEEtran}
\IEEEoverridecommandlockouts

\usepackage{siunitx}
\usepackage{amsfonts}
\usepackage{cite}
\usepackage{amsmath,amssymb,amsfonts}
\usepackage{algorithm}
\usepackage{graphicx}
\usepackage{textcomp}
\usepackage{xcolor}
\usepackage{algpseudocode}  
\usepackage{amsmath}

\usepackage{wrapfig}
\usepackage{subcaption}
\usepackage{balance} 
\AtBeginDocument{%
  \providecommand\BibTeX{{%
    Bib\TeX}}}
\usepackage{microtype}
\usepackage{graphicx}
\usepackage{booktabs} 
\usepackage{xcolor}
\usepackage{bm,bbm}
\usepackage{caption}
\usepackage[utf8]{inputenc} 
\usepackage[T1]{fontenc}    
\usepackage{hyperref}       
\usepackage{url}            
\usepackage{booktabs}       
\usepackage{amsfonts}       
\usepackage{nicefrac}       
\usepackage{microtype}      
\usepackage{color}  
\usepackage{amsmath, amsthm, multirow, paralist}
\usepackage{mathtools}
\usepackage{bm,bbm}
\usepackage{hyperref}
\hypersetup{colorlinks=true,breaklinks=true}

\usepackage{algorithm}
\usepackage{listings}

\newtheorem*{theorem*}{Theorem}

\def\BibTeX{{\rm B\kern-.05em{\sc i\kern-.025em b}\kern-.08em
    T\kern-.1667em\lower.7ex\hbox{E}\kern-.125emX}}

\makeatletter
\newcommand{\linebreakand}{%
  \end{@IEEEauthorhalign}
  \hfill\mbox{}\par
  \mbox{}\hfill\begin{@IEEEauthorhalign}
}
\makeatother

\begin{document}

\title{Less is more: Embracing sparsity and interpolation with Esiformer for time series forecasting}

\author{\IEEEauthorblockN{1\textsuperscript{st} Yangyang Guo\textsuperscript{*}}
\IEEEauthorblockA{\textit{Dept. Computer Science} \\
\textit{Xi'an Jiaotong University}\\
Xi'an, China \\
no.314016@stu.xjtu.edu.cn}
\and
\IEEEauthorblockN{2\textsuperscript{st} Yanjun Zhao\textsuperscript{*}}
\IEEEauthorblockA{\textit{Dept. Computer Science} \\
\textit{Xi'an Jiaotong University}\\
Xi'an, China  \\
yanjun.zhao@stu.xjtu.edu.cn}
\and
\IEEEauthorblockN{3\textsuperscript{rd} Sizhe Dang}
\IEEEauthorblockA{\textit{Dept. Computer Science} \\
\textit{Xi'an Jiaotong University}\\
Xi'an, China \\
darknight1118@stu.xjtu.edu.cn}
\linebreakand
\IEEEauthorblockN{4\textsuperscript{nd} Tian Zhou\textsuperscript{\textdagger}}
\IEEEauthorblockA{\textit{Damo Academy} \\
\textit{Alibaba Group}\\
Hangzhou, China \\
tian.zt@alibaba-inc.com}
\and

\IEEEauthorblockN{5\textsuperscript{th} Liang Sun}
\IEEEauthorblockA{\textit{Damo Academy} \\
\textit{Alibaba Group}\\
Bellevue, United States \\
liang.sun@alibaba-inc.com}
\and
\IEEEauthorblockN{6\textsuperscript{th} Yi Qian\textsuperscript{\textdagger}}
\IEEEauthorblockA{\textit{Dept. Computer Science} \\
\textit{Xi'an Jiaotong University}\\
Xi'an, China \\
yqian@mail.xjtu.edu.cn}

\thanks{\textsuperscript{*}Two authors contributed equally to this work.}
\thanks{\textsuperscript{\textdagger}Corresponding authors.}
}

\maketitle

\begin{abstract}
Time series forecasting has played a significant role in many practical fields. But time series data generated from real-world applications 
always exhibits high variance and lots of noise, which makes it difficult to capture
the inherent periodic patterns of the data, hurting the prediction accuracy significantly. To address this issue, we propose the Esiformer, which apply interpolation on the original data, decreasing the overall variance of the data and alleviating the influence of noise. 
What's more, we enhanced the vanilla transformer with a robust Sparse FFN. It can enhance the representation ability of the model effectively, and maintain the excellent robustness, avoiding the risk of overfitting compared with the vanilla implementation. Through evaluations on challenging real-world datasets, our method outperforms leading model PatchTST, reducing MSE by 
6.5\% and MAE by 
5.8\% in multivariate time series forecasting.  
Code is available at: https://github.com/yyg1282142265/Esiformer/tree/main.

\end{abstract}

\begin{IEEEkeywords}
Interpolation, Sparse Regression, Time Series forecasting
\end{IEEEkeywords}

\input{sections/1_introduction}

\input{sections/3_methods}
\input{sections/4_experiments}
\input{sections/5_conclusions}
\bibliographystyle{IEEEbib}
\bibliography{6_mybib}

\end{document}

%% file: sections/1_introduction.tex
\section{Introduction}\label{sec_intro}

Lots of real-world scenarios require  forecasting long sequence time-series, like weather forecasting, transportation planning, economics~\cite{Autoformer, FedFormer, zhou2023onefitsall, patchTST, Dlinear, KitaevKL20-reformer, liu2023itransformer, zhao2024sparse, hochreiter_long_1997_lstm,zhou2022film,GCformer}. Accurate long sequence time-series forecasting (LSTF) relies on a model's robust predictive ability, which includes accurately capturing the intricate periodic and trend patterns.

\begin{figure}[h]
\vskip -0.05in
\centering
\includegraphics[width=1\linewidth,trim=0 0 530 0,clip]{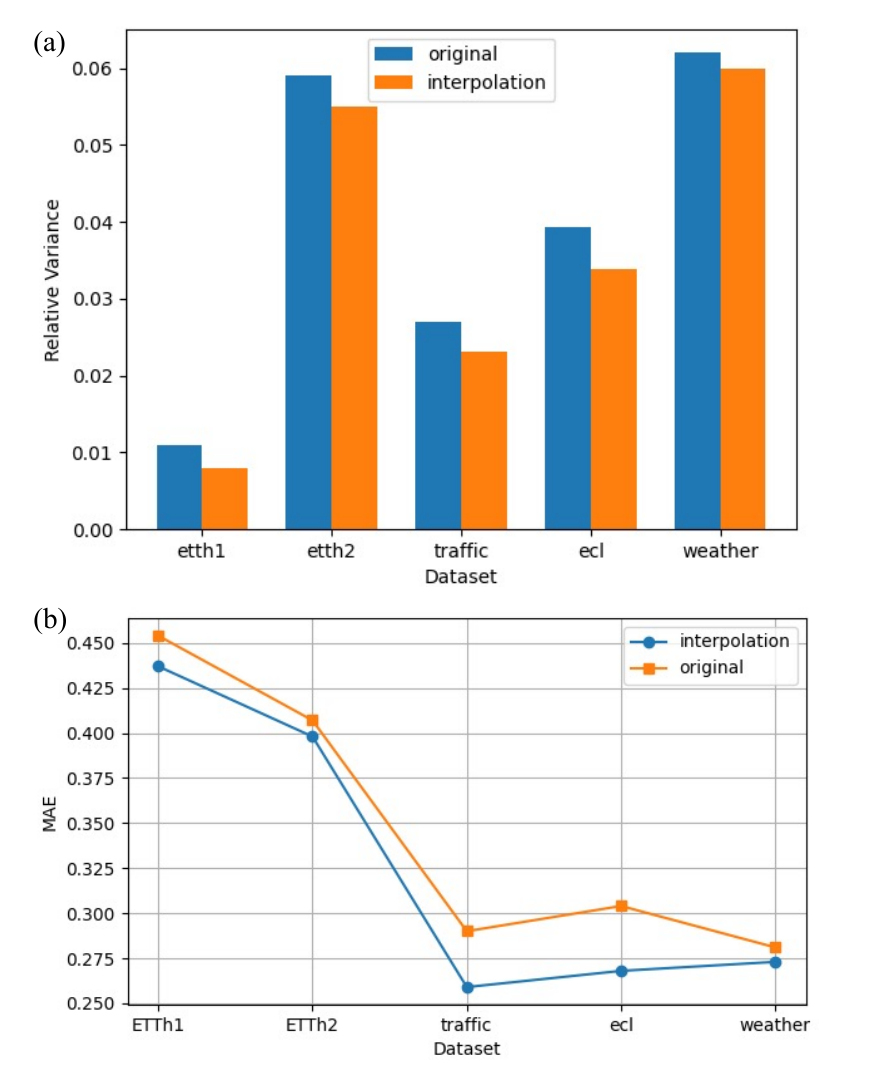}
\caption{Figure(a) compares the variance of five datasets before
and after the interpolation. Figure(b) shows the prediction
results on five datasets.}
\label{fig:intro}
\vskip -0.4in
\end{figure}



However, time series data collected from real-world applications are often replete with lots of noise~\cite{reversible, Non-stationary-Transformers,TIMESNET,wen2021robustperiod}, causing the high variance of the data, which severely impairs predictive performance. This is mainly because the data with high variance contains greater volatility and noise, which increases the degree of dispersion between data points, thereby masking the underlying trends and cyclical patterns. For forecasting models, this high volatility increases the difficulty of capturing the inherent patterns of the data, resulting in greater uncertainty in the forecast results.

\begin{figure*}[t]
\centering
\scalebox{0.75}{
\includegraphics[width=0.99\linewidth]{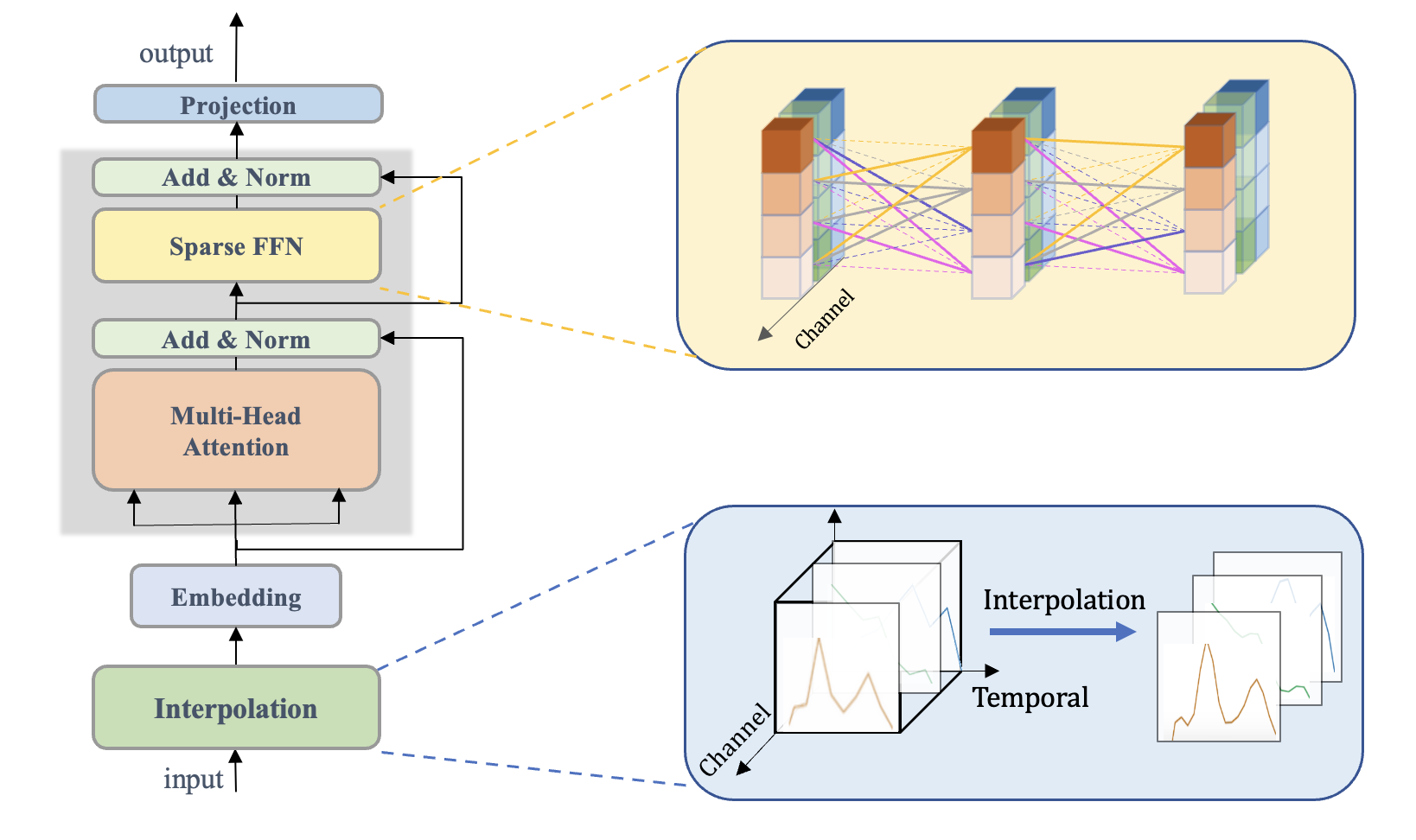}}
\caption{Framework of our proposed Esiformer.}
\label{fig:model_overview}
\vskip -0.25in
\end{figure*}

Many studies~\cite{interpolation,cheng2007reducing,vidal2016empirical} have suggested that interpolating data can reduce the data variance and thus enhance model performance. 
Similarly, we apply the interpolation technique to time series. Since all of the new interpolated data points are generated from the original data points, the content of information in data is preserved but the overall variance is reduced. Consequently, the model can capture temporal patterns more easily and obtain more accurate predictions.


Recently in NLP domains, transformers have made remarkable achievements~\cite{attention_is_all_you_need,kenton2019bert,zhang2022opt}, as demonstrated by applications such as ChatGPT~\cite{achiam2023gpt,floridi2020gpt}. These models imply that increasing the hidden dimension of the transformer can significantly enhance the representational capacity, leading to improved outcomes. So we consider whether we can apply a bigger hidden dimension in time series forecasting to get better prediction results. 
\input{tables_iTran/dime_reduction}
To investigate, we design two sets of experiments, using iTransformer models with different hidden dimension of the Feed Forward Network(FFN) for training. However, as shown in Table~\ref{tab:dimension_Reduction}, the experimental results show that increasing hidden dimensions (32 to 128) reduces training loss but barely boosts test performance, sometimes slightly hurting it. This suggests that the benefits in representational capacity achieved by expanding the hidden dimension are offset by the negative impacts of overfitting, resulting in no improvement in prediction performance. 
To address this issue, we propose the enhanced transformer, equipped with higher-dimensional sparse FFN. As a result, it can increase model representation capacity while effectively preventing overfitting.

To summarize, our contributions are listed as
follows:
\begin{enumerate}
\item The proposed Esiformer utilizes the interpolation to reduce the overall variance and mitigate the impact of noise, thereby improving the prediction accuracy.
\item We propose an enhanced transformer with a sparse FNN, replacing the original FFN. It embraces a larger hidden dimension to enhance the representation ability of the model while avoiding the overfitting, and it further deals with noise. 
\item We conducted lots of experiments to evaluate our Esiformer. Our empirical studies show that compared with PatchTST, Esiformer can reduce the prediction MAE by 
\textbf{5.8\%} and MSE by 
\textbf{6.5\%} for multivariate forecasting.
\end{enumerate}

%% file: tables_iTran/dime_reduction.tex
\vskip -0.1in
\begin{table}[h]
\centering

\caption{Univariate long-term series forecasting results of iTransformer with different FFN hidden dimensions for training. Input length $=96$ and prediction length $\in{96,192,336,720}$. Results are averaged from all prediction lengths. A lower Loss indicates better performance. }\vspace{-3mm}

\begin{center}
\begin{sc}
\scalebox{0.85}{
\begin{tabular}{c|cccccccccccccc}
\toprule
Methods &\multicolumn{2}{c|}{FFN-32} &\multicolumn{2}{c}{FFN-128}\\
\midrule
Metric  & trainloss & testloss & trainloss & testloss \\
\midrule
 etth1  & 0.348 & 0.076  & \textbf{0.344} & \textbf{0.074}\\
 etth2  & 0.376 & 0.193  & \textbf{0.373} & \textbf{0.192} \\
 electricity     & 0.333 & \textbf{0.314}   & \textbf{0.321} & 0.324\\
 traffic     & 0.140  & \textbf{0.142}    & \textbf{0.133} & \textbf{0.142} \\

\midrule
\bottomrule
\end{tabular}
\label{tab:dimension_Reduction}
}

\end{sc}
\end{center}
\vskip -0.15in
\end{table}

%% file: sections/3_methods.tex
\section{Method}\label{sec_methods}


We consider the following problem: given a collection of multivariate time series samples with lookback window $\textbf{X}: (x_1, ..., x_L) \in \mathbb{R}^{L \times M}$, we aim to forecast future values $\textbf{Y}:(x_{L+1}, ..., x_{L+T} ) \in \mathbb{R}^{T \times M}$. For convenience, we denote $X_{t,:}$ as the simultaneously recorded time points in step t, and
$X_{:,n}$ as the entire time series of each variate indexed by n.

\subsection{Model Structure}

      
      
      
      


Motivated by the aforementioned considerations, we introduce Esiformer, depicted in Figure \ref{fig:model_overview}. We interpolate the data for each channel independently.
\begin{gather*}
    X^{'}_{2L-1}  = Interpolation(X_{L}), \\
    H = Embedding(X^{'}_{2L-1}), \\
    H^{\prime} = TransformerBlock(H),  \\
    \hat{Y}_T = Projection(H^{\prime}), \\
\end{gather*}
\vspace{-20pt} 
\subsection{Interpolation}
In time series forecasting, time series with high variance tends to mask inherent patterns or trends in the data, making prediction more difficult. Previous work~\cite{interpolation,gallagher2008image,ferrara2012image} in computer vision has shown that interpolation on images causes a decrease in variance. For example, through the bilinear interpolation, the variance of interpolated pixels is reduced to $1/4$ of that of the original pixels. 

Thus, we propose to interpolate the time series data to reduce the variance of data, as shown in Figure~\ref{fig:model_overview}, and thereby facilitate prediction\cite{onder2014forecasting}. Here we show four interpolation methods.





\subsubsection{TwoAver}


\[  
x'_i =   
\begin{cases}  
x_{\lceil i/2 \rceil} & \text{if } i \text{ is odd} \\  
\frac{x_{\lceil i/2 \rceil} + x_{\lceil i/2 \rceil + 1}}{2} & \text{otherwise}  
\end{cases}  
\] 

\subsubsection{FourAver}

\begin{align*}  
&x'_{3i} = \frac{x_{2i-1} + x_{2i} + x_{2i+1} + x_{2i+2}}{4} \\  
&x'_{3i-1} = x_{2i} \\  
&x'_{3i-2} = x_{2i-1}  
\end{align*} 

\subsubsection{Spline}
Spline aims to find a function $s(x)$ that is a cubic polynomial on each interval $[x_i,x_{i+1}]$ and satisfies the following conditions:

Interpolation condition: 
\[
s(x_i) = y_i \quad \text{for all} \quad i = 1, 2, \ldots, n
\]

Continuity condition: 
\[
s_{i-1}(x_i) = s_i(x_i) \quad \text{for all} \quad i = 2, 3, \ldots, n
\]
where $s_{i-1}(x)$ and $s_i(x)$ are cubic polynomials on the intervals $[x_{i-1}, x_i]$ and $[x_i, x_{i+1}]$, respectively.

Smoothness condition: 
\[
s_{i-1}'(x_i) = s_i'(x_i) \quad \text{and} \quad s_{i-1}''(x_i) = s_i''(x_i) \quad \text{for all } i
\]

The interpolated "sequence" is not actually a set of discrete points, but rather a smooth curve s(x). 
So we calculate the value of s(x) at a fixed interval between the original data points to generate a new, denser sequence.


  
  

\subsubsection{Rbf}
RBF tries to find a function $s(x)$ that satisfies $s(x_i) = y_i$ for all $i = 1, 2, \ldots, n$, which can usually be expressed as:
\[
s(x) = \sum_{i=1}^{n} \lambda_i \phi(\|x - x_i\|)
\]
Wherein, $\phi(\|x - x_i\|)$ is the radial basis function, which depends only on the Euclidean distance $\|x - x_i\|$ between $x$ and $x_i$, and $\lambda_i$ is the coefficient to be determined. For the radial basis function $\phi$, we use Gaussian function, defined as:
\[
\phi(r) = \exp\left(-\frac{r^2}{2\sigma^2}\right), 
\]
where $r = \|x - x_i\|$ is the distance, $\sigma$ is the parameter that controls the width of the function.





\subsection{Sparse FFN}



The standard linear regression model postulates a linear relationship between the dependent variable y and the independent variables X, formulated as $ y = X \beta + \epsilon $. Least squares optimization is used to find $\beta$ to best fit observed $y$ values.
\begin{equation}
\min_{\beta \in \mathbb{R}^n}g(\textbf{y}-\textbf{X}\beta),
\label{eq:standard linear regression}
\end{equation}

where $\textbf{y} \in \mathbb{R}^m $ and $ \textbf{X} \in \mathbb{R}^{m \times n}$ are data and $g$ is some convex function, typically a norm.

To prevent overfitting, regularization(sparsity) is employed in model parameter estimation:
\begin{equation}
\min_{\beta}g(\textbf{y}-\textbf{X}\beta)+h(\beta),
\label{eq:sparse linear regression}
\end{equation}

Regularization via a convex penalty $h$ simplifies models, focuses on key features, and boosts generalization by reducing overfitting to noise.

Therefore, after self-attention mechanism in EncoderBlock, we choose sparse linear network for temporal modeling to improve the generalization ability of the model. 

In regression analysis, data is ideally noise-free. However, practical data often contains errors and adversarial noises, impacting the accuracy and reliability of  models \cite{singh1999noise,flores2016comparison,sangiorgio2021forecasting,soofi2002nonlinear}. Neglecting these factors may lead models to learn incorrect patterns. Compared to traditional regression, robust regression explicitly considers the noise in the data when modeling. This approach aims to improve the stability and robustness of the model in a noisy environment. Specifically, we define an uncertainty set to include possible noise or perturbations and minimize the worst-case loss on this uncertainty set. As in \cite{el1997robust,lewis2010lipschitz,ben2009robust,xu2008robust}, this approach may take the form:
\begin{equation} 
\min_{\beta}{\max_{\Delta \in \psi }g(y-(X+\Delta)\beta)},
\label{eq:robust linear regression}
\end{equation}
where the set $\psi \subseteq \mathbb{R}^{m \times n}$ characterizes the user’s belief about uncertainty on the data matrix X.


Sparse regression focuses on the sparsity of model parameters, that is, most parameters are zero or close to zero, which helps the model's interpretability and generalization ability. Robust regression focuses on outliers or noise in the data, aiming to find a regression model that is insensitive to outliers. Although they deal with different problems in essence, sparse regression (\ref{eq:sparse linear regression}) and robust regression (\ref{eq:robust linear regression}) are equivalent if $g$ is a semi-norm and $h$ is a norm~\cite{sparse_robust}. Furthermore, \cite{burgard2020generalized} prove that they are equivalent when $g$ and $h$ are functions of semi-norms and norms. They also show the equivalence between robust estimation in linear mixed models and comprehensively penalized regularized parameter estimation. So sparse FFN can positively impact robustness of the model while simultaneously preventing overfitting. With interpolation, the model can better adapt to noisy environments.



\input{tables_iTran/MultiVariate}

%% file: tables_iTran/MultiVariate.tex
\begin{table*}[t]
\centering
\begin{sc}

\caption{Multivariate long-term series forecasting results. All the results are with same input length $=96$ and averaged from 4 different prediction lengths $\in \{96,192,336,720\}$ . A lower MAE/MSE indicates better performance.The best results are highlighted in bold and the second best are underlined. 
}
\scalebox{0.85}{
\begin{tabular}{cc|cccccccccccccccccc}

\toprule
&{Methods} &\multicolumn{2}{c|}{Esiformer} &\multicolumn{2}{c|}{PatchTST}  &\multicolumn{2}{c|}{Dlinear}  &\multicolumn{2}{c|}{FEDformer} &\multicolumn{2}{c|}{Autoformer} &\multicolumn{2}{c|}{TimesNet} &\multicolumn{2}{c|}{SCINet} &\multicolumn{2}{c}{Crossformer}\\
\midrule
&{Metric} & MSE  & MAE  & MSE  & MAE & MSE  & MAE& MSE  & MAE& MSE  & MAE& MSE  & MAE & MSE  & MAE & MSE & MAE\\

\midrule
&{ETTh1}
 & \underline{0.443} & \textbf{0.437}  & 0.469 & 0.454 & 0.456 & 0.452 & \textbf{0.440} & 0.460 & 0.496 & 0.487 & 0.458 & \underline{0.450} & 0.747 & 0.647 & 0.529  & 0.522 \\

\midrule
&{ETTh2}
 & \textbf{0.376} & \textbf{0.398}  & \underline{0.387} & \underline{0.407} & 0.559 & 0.515 & 0.437 & 0.449 & 0.450 & 0.459 & 0.414 & 0.427 & 0.954 & 0.723 & 0.942  & 0.684 \\

\midrule
&{Electricity}
 &\textbf{ 0.175} & \textbf{0.259}  & 0.205 & \underline{0.290} & 0.212 & 0.300 & 0.214 & 0.327 & 0.227 & 0.338 & \underline{0.192} & 0.295 & 0.268 & 0.365 & 0.244 & 0.334 \\

\midrule
&{Traffic}
 & \textbf{0.433} & \textbf{0.268}  & \underline{0.481} & \underline{0.304} & 0.625 & 0.383 & 0.610 & 0.376 & 0.628 & 0.379 & 0.620 & 0.336 & 0.804 & 0.509 & 0.550 & 0.304 \\

\midrule
&{Weather}
 &\textbf{0.257} & \textbf{0.273}  & \underline{0.259} & \underline{0.281} & 0.265 & 0.317   & 0.309 & 0.360  & 0.338 & 0.382 & 0.259 & 0.287 & 0.292 & 0.363 & 0.259 & 0.315 \\

\midrule

\bottomrule
\end{tabular}
\label{tab:MultiVariate}
}
\vskip -0.15in
\end{sc}
\end{table*}

%% file: sections/4_experiments.tex
\section{Experiments}
\label{sec_experiments}

\subsection{Dataset and implementation details}

We conduct comprehensive experiments on widely used datasets and choose the renowned PatchTST as our baseline and follow the same experimental setup. All forecasting results are run on GeForce GTX 1080 Ti GPU.




\subsection{Long-term Forecasting}


The multivariate time series forecasting results are listed in Table \ref{tab:MultiVariate}, where the best methods are highlighted in bold. Our Esiformer shows superior forecasting performance compared to the recent SOTA methods. Remarkably, this model demonstrates notable performance advantages over PatchTST when handling the high-dimensional time series datasets. Quantitatively, Esiformer achieves an overall \textbf{10.0\%} reduction on MSE and \textbf{11.8\%} reduction on MAE for Traffic dataset, which indicates that our model can effectively handle real-world time series forecasting.


%

\subsection{Ablations}

\textbf{Interpolation} We study the effects of interpolation in Table \ref{tab:Interpolation_Ablation}. To verify the effectiveness of interpolation, we use different interpolation methods to perform univariate forecasting on the 
traffic dataset, and the input time series length is fixed to 96. And in order to eliminate the impact of the sparse mechanism, this part of the ablation experiment excludes it. The results show that interpolating the original time series can effectively reduce the variance and thus improve the forecasting accuracy. Compared with no interpolation, interpolation reduces MSE by 
\textbf{10.3\%} on average and 
\textbf{7.8\%} for MAE.

\input{tables_iTran/ablation_inter_traffic}
In practice, cubic spline interpolation and RBF interpolation tend to generate smoother curves, which is conducive to capturing long-term trends in the data and thus performing better in long-term predictions. However, in medium- and short-term predictions, they distort important local features and are therefore inferior to the other two simple interpolation methods. In addition, the effect of RBF interpolation depends on parameter selection (such as basis function shape and width). If the parameters are not selected properly, the interpolation effect may not be ideal. In contrast, simple interpolation methods are more reliable and easy to optimize in short-term predictions due to their directness and controllability.

\textbf{Sparse FFN} As mentioned above, when hidden\_dimension increases, although the representation power of model improves, the risk of overfitting also rises, with these two effects balancing each other to some extent. So we want to introduce sparse FFN to alleviate overfitting. To validate the effectiveness of the sparse mechanism, we present detailed ablation study results, summarized in Table \ref{tab:Sparse_Ablation}. Similarly, this part of the ablation experiment excludes the interpolation mechanism. 
The results show that incorporating the sparse mechanism led to significant improvements in MSE and MAE for both electricity and traffic datasets. The introduction of the sparse mechanism improves the MSE by \textbf{8.97\%} and the MAE by \textbf{5.49\%} in the electricity dataset prediction task. And for the traffic dataset, the MSE is improved by \textbf{6.65\%} and the MAE is improved by \textbf{6.60\%}. 

We also combine our method with FEDformer~\cite{FedFormer} and Informer~\cite{haoyietal-informer-2021} as a plug-in to evaluate the generality of
our Esiformer. As shown in Table~\ref{tab:boosting}, our method can yield a boosting result and enhance the prediction accuracy.

\input{tables_iTran/Sparse}

\input{tables_iTran/boosting}

%% file: tables_iTran/ablation_inter_traffic.tex
\begin{table}[h]
\centering
\begin{sc}

\caption{Interpolation Ablation: Univariate long-term series forecasting results on Traffic with input length $L = 96$ and prediction length $T \in \{96, 192, 336, 720\}$. 
}
\vspace{-2mm}
\scalebox{0.68}{
\begin{tabular}{cc|cccccccc}

\toprule
&{Methods} &\multicolumn{2}{c|}{96} &\multicolumn{2}{c|}{192} &\multicolumn{2}{c|}{336}  &\multicolumn{2}{c}{720} \\

\midrule
&{Metric} & MSE  & MAE & MSE & MAE & MSE  & MAE & MSE  & MAE \\
\midrule

&{Spline} & 0.143           & 0.222           & 0.222      & 0.314      & 0.176       & 0.271      & 0.316      & 0.392       \\

\midrule

&{TwoAver}  & 0.142           & 0.222           & 0.309      & 0.382     & 0.225       & 0.322      & 0.313      & 0.391\\

\midrule

&{FourAver} & 0.143           & 0.223           & 0.183       & 0.274      & 0.172       & 0.266      & 0.313      & 0.393  \\

\midrule
&{Rbf} &0.143          & 0.222        & 0.183       & 0.275      & 0.185       & 0.281      & 0.316      & 0.391 \\

\midrule
&{No Interpolation} & 0.144           & 0.223           & 0.255       & 0.343      & 0.237       & 0.331     & 0.343      & 0.425 \\

\midrule

\bottomrule
\end{tabular}
\label{tab:Interpolation_Ablation}
}
\vskip -0.10in
\end{sc}
\end{table}

%% file: tables_iTran/Sparse.tex
\begin{table}[t]
\centering
\begin{sc}
\caption{Ablation experiment for Sparse FFN: Univariate long-term series forecasting results on four different datasets with input length $L=96$.}
\vspace{-1mm}
\scalebox{0.68}{

\begin{tabular}{c|c|cccccccccccccccc}

\toprule
\multicolumn{2}{c|}{Pred\_len}
&\multicolumn{2}{c|}{96}&\multicolumn{2}{c|}{192}&\multicolumn{2}{c|}{336}&\multicolumn{2}{c}{720}\\
\midrule

\multicolumn{2}{c|}{Metric}  & MSE & MAE  & MSE & MAE & MSE & MAE  & MSE & MAE \\

\midrule

\multirow{2}{*}{etth2} 
&Sparse FFN & 0.138 & 0.288 & 0.183 & 0.337 & 0.213 & 0.364 & 0.223 & 0.378  \\
&original  & 0.145 & 0.297 & 0.187 & 0.341 & 0.213 & 0.365 & 0.227 & 0.382 \\

\midrule

\multirow{2}{*}{etth1} 
&Sparse FFN & 0.060 & 0.186 & 0.074 & 0.207 & 0.085 & 0.226 & 0.085 & 0.230  \\
&original & 0.061 & 0.188 & 0.077 & 0.212 & 0.085 & 0.226 & 0.086 & 0.232 \\

\midrule

\multirow{2}{*}{electricity} 
&Sparse FFN & 0.261 & 0.369 & 0.298 & 0.387 & 0.358 & 0.423 & 0.402 & 0.457 \\
&original & 0.344 & 0.427 & 0.307 & 0.393 & 0.398 & 0.454 & 0.400 & 0.457 \\

\midrule

\multirow{2}{*}{traffic}
&Sparse FFN & 0.148 & 0.228 & 0.139 & 0.213 & 0.137 & 0.216 & 0.152 & 0.235          \\
&original & 0.171 & 0.264 & 0.142 & 0.215 & 0.147 & 0.235 & 0.157 & 0.241    \\

\midrule
\bottomrule

\end{tabular}

}

\label{tab:Sparse_Ablation}
\end{sc}
\end{table}

%% file: tables_iTran/boosting.tex
\begin{table}[t]
\centering
\begin{sc}
\caption{Boosting results on FEDformer and Informer.}
\scalebox{0.68}{

\begin{tabular}{c|c|cccccccccccccccc}

\toprule

\multicolumn{2}{c|}{Methods}  &FEDformer &FED\_enhanced &Informer  &Inf\_enhanced \\

\midrule

\multirow{3}{*}{Electricity} 
&96 & 0.253/0.370 & 0.244/0.363 & 0.258/0.367 & 0.259/0.363\\
&192  & 0.282/0.386 & 0.281/0.386 & 0.285/0.388 & 0.281/0.377\\
&336  & 0.346/0.431 & 0.342/0.429 & 0.336/0.423 & 0.336/0.423\\
\midrule

\multirow{3}{*}{Weather} 
&96 & 0.0062/0.0620 & 0.0035/0.0490& 0.0040/0.0440 & 0.0033/0.0424\\
&192 &    0.0060/0.0620       &  0.0045/0.0540 & 0.0020/0.0400 & 0.0028/0.0400\\
&336 &    0.0041/0.0500      &  0.0056/0.0600 &0.0040/0.0490 & 0.0029/0.0407\\
\midrule

\bottomrule

\end{tabular}

}

\label{tab:boosting}
\end{sc}
\vskip -0.15in
\end{table}

%% file: sections/5_conclusions.tex
\section{Conclusion}
\label{sec_conclusion}
This paper introduces a Transformer-based time series forecasting model Esiformer. It interpolates the original data, reduces the overall variance of the data and mitigates the impact of noise. In addition, we use robust sparse FFN to enhance the vanilla Transformer. The sparsification strategy not only reduces the number of parameters and model complexity, but also enhances the locality of features through sparse connection patterns, which helps the model to better learn important features while reducing the interference of noise information. It can effectively enhance the representation ability of the model and avoid the risk of overfitting. Experiments show that our proposed model outperforms PatchTST on five mainstream time series forecasting datasets.

\clearpage